\documentclass[12pt]{article}

\usepackage{graphicx}
\usepackage{enumitem}
\usepackage{varwidth}
\usepackage{multicol}
\usepackage[numbers,sort&compress]{natbib}
\usepackage{url}

\begin{document}

\title{Taxonomy and Modular Tool System\\for Versatile and Effective\\Non-Prehensile Manipulations}

\author{Cedric-Pascal Sommer, Robert J.~Wood, and~Justin Werfel}

\date{June 25, 2025}

\maketitle

\begin{abstract}
General-purpose robotic end-effectors of limited complexity, like the parallel-jaw gripper, are appealing for their balance of simplicity and effectiveness in a wide range of manipulation tasks.
However, while many such manipulators offer versatility in grasp-like interactions, they are not optimized for non-prehensile actions like pressing, rubbing, or scraping---manipulations needed for many common tasks. To perform such tasks, humans use a range of different body parts or tools with different rigidity, friction, etc.\ according to the properties most effective for a given task.
Here, we discuss a taxonomy for the key properties of a non-actuated end-effector, laying the groundwork for a systematic understanding of the affordances of non-prehensile manipulators.
We then present a modular tool system, based on the taxonomy, that can be used by a standard two-fingered gripper to extend its versatility and effectiveness in performing such actions.
We demonstrate the application of the tool system in aerospace and household scenarios that require a range of non-prehensile and prehensile manipulations.

\end{abstract}

\let\originaltextit\textit
\renewcommand{\textit}[1]{%
  {\fontsize{6}{1}\selectfont\originaltextit{#1}}%
}

\section{Introduction}

While most research in robotic manipulation focuses on grasping, non-prehensile manipulation\footnote{Here, we use the term ``non-prehensile'' to refer to interactions where the end-effector does not fully enclose, encage, or encapsulate the manipulated object.}\citep{cutkosky1989grasp}
has received comparatively less attention, particularly for tasks that are not highly dynamic. However, non-prehensile interactions are ubiquitous,  essential for a wide range of task classes, and potentially advantageous in multiple respects over prehensile manipulations \citep{lynch1998issues,menon2018non,lynch1996stable}.

In non-prehensile manipulations, the geometric and material properties of the end-effector can be critical to task performance if motion planning and control should be kept at reasonable complexity, and choosing an end-effector with the appropriate properties is both essential and routine. 
For instance, a lottery scratch card cover is scratched off using a fingernail or coin, an adhesive sticker is pressed onto a surface using the palm, and a precise line is traced on a touch screen using a stylus. 
Moreover, note that many distinct non-prehensile tasks can be executed with a similar, simple motion by the arm, yet produce widely different interaction results due to differences in their end-effector properties.
Consider the different task results produced by a similar motion according to whether the end-effector is an ice scraper, paint roller, or push broom.

This observation points to a key consideration for the design of a general-purpose manipulation system: to appropriately handle the range of operations that everyday tasks require, such a system should have end-effectors with a corresponding range of properties available to it. 

\begin{figure*}[tb]
    \centering
    \includegraphics[width=\textwidth,height=\textheight,keepaspectratio]{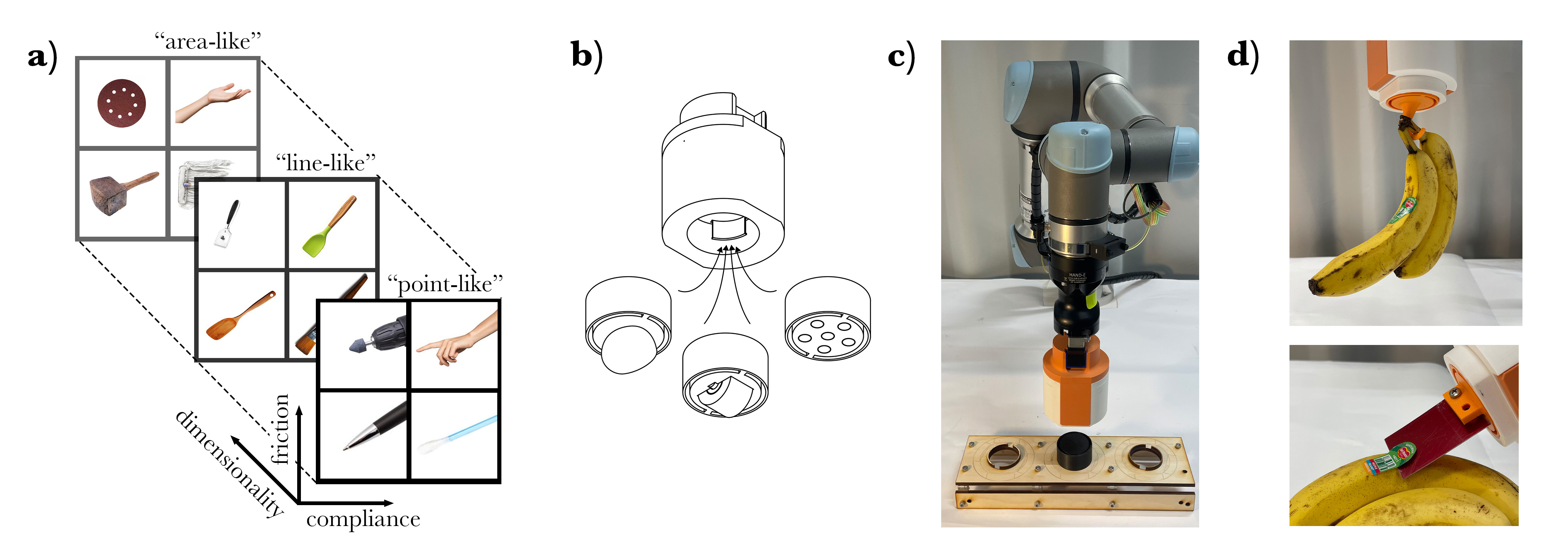}
    \caption{Ably executing a range of different non-prehensile tasks requires a corresponding range of end-effector properties. a) A formal characterization scheme for such properties aids both tool design and selection. b) A modular tool system with interchangeable, passive end-effectors can provide an autonomous robot with the required range for different tasks. c) The system can be autonomously operated, and different end-effectors mounted, by a standard robot arm with a two-fingered gripper. d) Examples of two end-effectors that facilitate different types of tasks. Top: a hook aids with lifting a bunch of bananas; bottom: a flexible peel aids with removing a sticker from a banana.}
    \label{fig:abstract}
\end{figure*}

This principle is evident in natural manipulators, like the human hand, which features different structures that can be used according to the needs of a given task. For instance, these end-effectors can be small (fingertip), medium (finger), or large (palm); rigid (knuckles of a fist) or soft (heel); sharp (nail edge) or blunt (finger pad); smooth (nail surface) or grippy (finger pad ridges); and so on.
Analogous versatility for artificial manipulators can be provided through modular systems of interchangeable end-effectors, as with electric drill attachments ranging from mixing heads to buffing wheels, or active tool changer systems, as with those found in CNC machining \citep{hollis1968automatic, ryuh2006automatic, gordon2016robotic} and industrial robots \citep{silvers1986tool}.

If autonomous robots are expected to perform a wide range of common tasks, a system is needed to provide the corresponding range of properties. Most manipulation studies, focusing on grasping, use a single finger surface material and geometry for all considered tasks \citep{marwan2021comprehensive}. Prior studies concerning non-prehensile manipulations are usually focused on a single, typically highly dynamic task, with a correspondingly specialized end-effector \citep{mason1986mechanics,lynch1996stable,yu2012research,batz2010dynamic}. Existing modular systems, like for drill attachments, require human intervention to change end-effectors \citep{dewalt_dcd793d1,milwaukee_2904_20}, and active modular tool-changer systems for autonomous operation are often expensive and bulky \citep{ati_robot_tool_changer,robotiq_tool_changer,ryuh2006automatic,smith2020modular}. A simple, inexpensive system allowing autonomous switching between a range of end-effector properties, adaptable to different robot platforms, could provide a widely adoptable tool facilitating the autonomous execution of a variety of non-prehensile tasks.

As a first step in developing such a tool, a need exists for a systematic characterization of the spectrum of end-effector properties relevant to non-prehensile interactions. Existing manipulation taxonomies used in robotics---focused on grasping, and not concerned with the mechanical properties of the gripper---typically exclude non-prehensile manipulations altogether \citep{feix2015grasp,cutkosky1989grasp}, or group all non-prehensile manipulations into a few high-level categories \citep{bullock2011classifying,mason1986mechanics} that focus on the \emph{type of interaction} rather than the \emph{properties of the interactor}. A taxonomy for non-prehensile interactions would not only inform the design of a tool system to facilitate such actions, but would provide a principled framework for the selection of an appropriate tool set for a given task.

In this paper, we introduce a novel taxonomy for non-prehensile manipulations (\S\ref{sec:characterization}), and, building on this classification scheme, design a modular tool system that can be used by a standard autonomous robotic arm with a two-finger gripper (\S\ref{sec:toolsystem}) (Fig.~\ref{fig:abstract}). The system comprises a single passive tool holder and a palette of passive tool inserts providing a range of non-prehensile affordances identified in the characterization scheme. 
We show empirically that the reliability of our prototype tool changer system is in line with comparable existing tool changers, and demonstrate its use in example scenarios drawn from space and household domains, each demanding a different set of non-prehensile manipulation classes (\S\ref{sec:casestudies}). 

\section{Related Work}
\subsection{Non-prehensile manipulation materials}

The majority of robotic manipulation research is concerned with prehensile manipulation; tasks in the non-prehensile domain have received less attention. 
It has been argued that this discrepancy arises from the fact that prehensile control systems are less complex once the object is grasped \citep{ruggiero2018nonprehensile}.
In comparison, non-prehensile manipulations are underactuated \citep{mason1999progress} and exhibit open force linkages \citep{serra2016robot}, leading to non-linearity because of interactions between the end-effector, the object, and the task environment \citep{feix2014analysis}. 

Nevertheless, non-prehensile interactions are essential for various actions, like pushing, throwing, batting, balancing, etc.\ \citep{menon2018non}. When a choice between a prehensile and non-prehensile manipulation can be made, the latter can offer advantages, like increased operational space, minimized execution time, higher dexterity \citep{menon2018non}, ability to interact with objects not suited for available grippers to handle \cite{lynch1996stable}, and opportunities to exploit features in the task environment to improve grasp performance \citep{eppner2015planning}.

Most prior work in the non-prehensile domain focuses on controllers in highly-dynamic niche scenarios (e.g., playing billiards \citep{shi2017dynamic}, ping-pong \citep{yu2012research}, catching balls \citep{riley2002robot, batz2010dynamic}, or spinning a flower stick \citep{aoyama2015realization}). As a result, the end-effectors used in controller-heavy studies are usually highly-specialized, single-purpose manipulators (in those examples, a cue, paddle, bowl, and rod, respectively); their shape and surface material is not intended to generalize to other environments.

Some works that are less concerned with non-prehensile control studies show greater versatility in the shape of the end-effector to enable task success. Examples can be found in hooks (e.g., holding a bag handle) \citep{menon2018non}, pushing/sliding manipulators (e.g., closing a door) \citep{prats2010reliable}, rolling (e.g., rolling a ball along a beam) \citep{virseda2004modeling}, and palmar interaction end-effectors (e.g., carrying a tray) \citep{huang2001nonprehensile}.

Earlier work in parts-orienting tasks similarly focuses on the object motion and does not consider how the properties of the interface between the end-effector and object can affect task success
\citep{mason1982manipulator, mason1986mechanics, grossman1975orienting, erdmann1988exploration, grossman1975orienting, lynch1996stable, mason1993dynamic}.

\subsection{Prehensile manipulation materials}

Some studies of prehensile manipulation have explored using different contact materials to increase a gripper's affordances in specific contexts, e.g., smooth aluminum for decluttering scenarios \citep{imtiaz2023prehensile, dogar2012planning}, balsa wood and copper alloy for improved textile handling \citep{ono1991robot}, and compliant materials \citep{shimoga1992soft} to adapt to the shape of unknown objects \citep{eppner2013grasping}, attenuate impact force \citep{biagiotti2003mechatronic}, and improve grasp reliability with uneven objects \citep{lotti2002mechanical}. In addition to varying a manipulator's surface materials, some researchers investigate the use of sub-surface soft layers in robotic fingers for even better object conformability \citep{lotti2002novel} and anthropomorphicity \citep{biagiotti2004far}. Further grasp affordances can be made available to a gripper by exploiting variations in the geometry and materials of environmental constraints \citep{eppner2015exploitation}.

Some studies combine multiple distinct materials or geometries in one manipulator, making two or three options available to extend a robot's ability to perform multiple tasks requiring different properties (e.g., adding fingernails \citep{murakami2004novel}, or fingers that can be flipped in orientation \citep{watanabe2021variable}). A few incorporate additional actuators to rotate fingers around their axes to expose different materials to a grasped object \citep{morita2000human, yoshimi2012picking}, or to provide active finger surfaces featuring miniature conveyor-belts to modulate friction selectively \citep{tincani2012velvet, ma2016hand}.

\subsection{Passive prosthetics}

Another domain concerned with manipulator properties is upper limb prosthetics, whose goal is to enable their wearer to carry out common manipulation tasks. Passive cosmetic prostheses, with no actuation, usually offer a small number of distinct contact surfaces and materials (e.g., soft foam padding or a rubbery finger texture \citep{plettenburg2009wilmer, maat2018passive}, mimicking the different hand surface materials humans use for different tasks. 
Wearers learn to selectively use different surfaces in different contexts \citep{pillet1983esthetic, crandall2002pediatric}, again pointing to the importance of having different manipulator surface materials available for non-prehensile interactions.

Like their natural counterparts, passive prosthetics can only offer so many distinct surface materials. To further extend the versatility of affordances, modular prostheses enable users to switch out different tool inserts depending on the task \citep{webster2001sports}. Consequently, such prostheses can cover a wider range of tasks than any single passive prosthetic. The inserts offer variation in contact shape and materials, aiding tasks like playing baseball \citep{truong1986baseball}, rowing \citep{highsmith2009design, radocy1987upper, radocy1992special}, construction and woodworking \citep{britishpathe_blackrock_1922}, or playing musical instruments \citep{norris1993applied}. The idea for such modular prostheses is also popular in speculative fiction \citep{dick1965three, hook1991}.

\subsection{Robotic tool changer systems}

Tool changers for autonomous robots couple a
single generic tool mount with specialized tools, each delivering a desired capability to the system. 
Most tool changers are designed for high-precision, high-rigidity machining and manufacturing tasks, like CNC manufacturing \citep{hollis1968automatic, ryuh2006automatic, gordon2016robotic, rogelio2014development} and industrial robots \citep{ati_robot_tool_changer, robotiq_tool_changer, schunk_tool_changer, meghdari2000design, li2022modular}, and are accordingly bulky, expensive, and specialized. Some research has considered simpler, passive tool changers under low-weight, low-volume, and low-complexity constraints for mobile robots \citep{gyimothy2011experimental}, designed for specific contexts, such as in agricultural settings \citep{berenstein2018open}, nuclear waste management \citep{pettinger2019passive}, and aerospace applications \citep{li2022modular}. 
In some cases, a passive tool changer may accommodate active, powered inserts \citep{iqbal2021detachable, salvietti2018co, li2022modular} (with corresponding costs compared to passive ones).

\subsection{Manipulation taxonomies}

Existing grasp taxonomies classify poses and characteristics of a manipulator, typically focusing on the human hand: the anatomy of the hand \citep{schlesinger1919mechanische}, the muscle groups involved in a grasp \citep{stival2019quantitative}, the number of fingers involved \citep{napier1956prehensile}, in-hand movements \citep{napier1956prehensile, landsmeer1962power}, the overall geometry or shape of a grasp \citep{cutkosky1989grasp, slocum1946disability} or the object it is interacting with \citep{blanco2024t,sun2022multi}, or a combination of these dimensions \citep{bullock2011classifying,feix2015grasp,krebs2022bimanual}. Some grasp taxonomies have also been presented to characterize individual classes of robotic manipulators \citep{mehrkish2021comprehensive}.
Importantly, most existing taxonomies exclude non-prehensile dimensions and properties of the interaction between the manipulator and the manipulated object \citep{feix2015grasp}.
Even in the rare cases where the author acknowledges the importance of materials like the hard fingernail or the soft finger pulp \citep{kapandji1982physiology}, this dimension is ultimately excluded from the taxonomy. 
Taxonomies that do take non-prehensile considerations into account still lack differentiation by mechanical properties.
Consequently, end-effectors that present a similar shape, e.g., a hammer face and a palm held flat, would appear in the same category \citep{bullock2011classifying} despite their lack of interchangeability for most tasks.
Therefore, there is a need for a classification system that captures these additional factors.

\section{Taxonomy\label{sec:characterization}}

\begin{table}[tp]
    \centering
    \includegraphics[width=\columnwidth,height=\textheight,keepaspectratio]{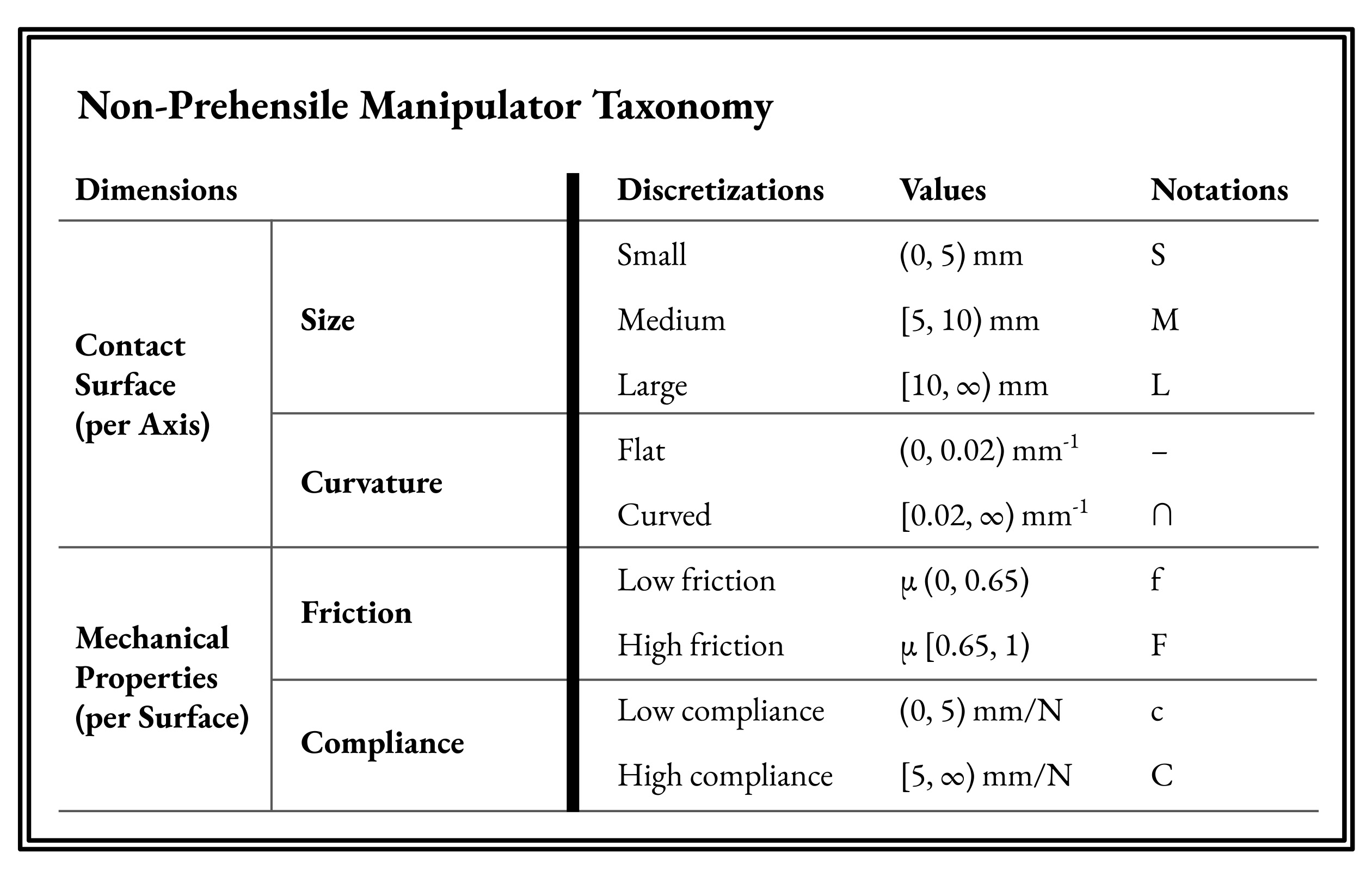}
    \caption{Taxonomy for end-effector properties relevant for non-prehensile manipulation. Left: Continuous dimensions characterizing a manipulator surface. Right: Example of a specific choice of discretization.}
    \label{fig:characterization_table}
\end{table}

\begin{figure*}[tp]
    \centering
    \includegraphics[width=\textwidth, height=0.9\textheight, keepaspectratio]{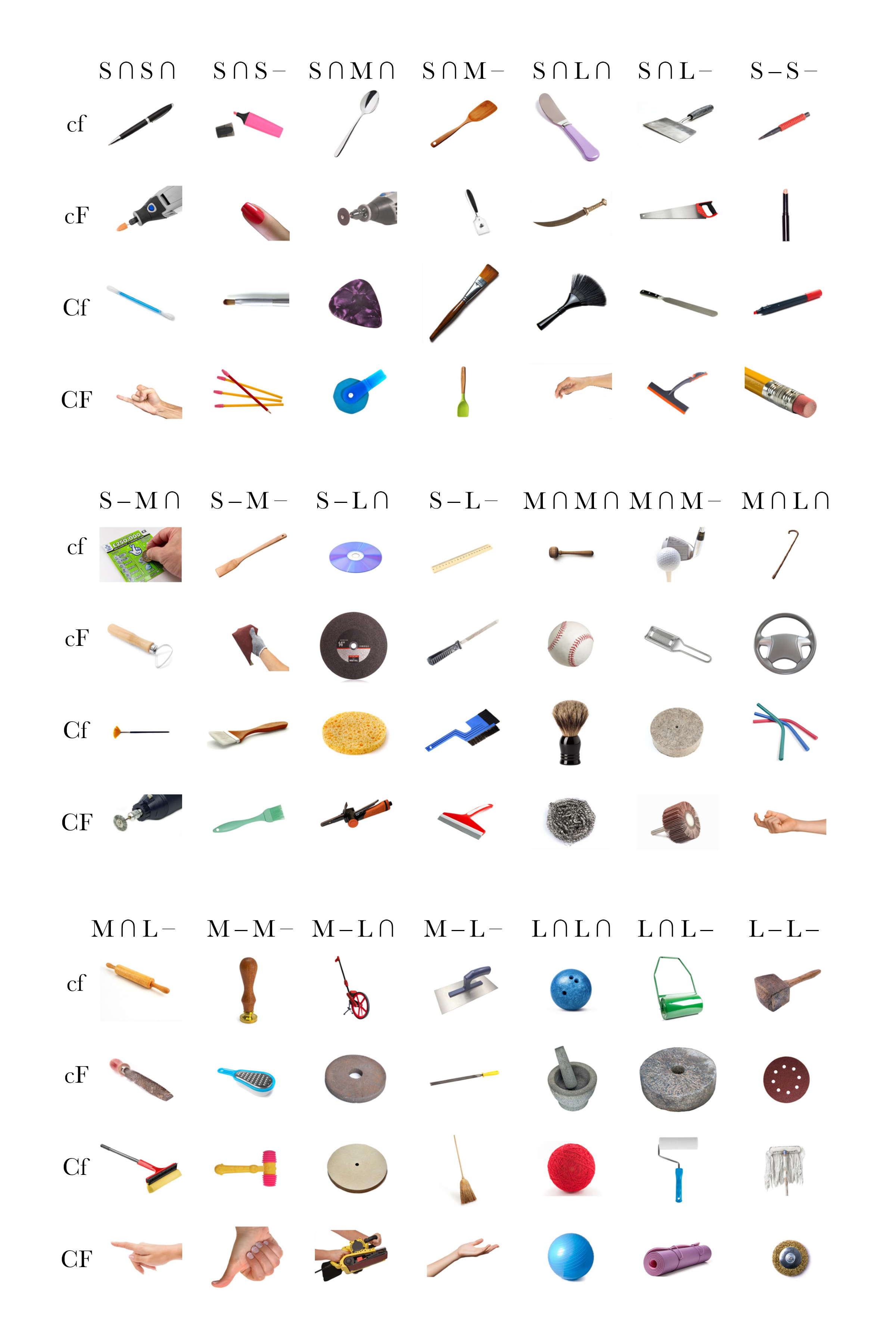}
    \caption{Exhaustive set of examples of all possible classes using the discretization provided on the right-hand side of Table~\ref{fig:characterization_table} (see Fig.~\ref{fig:appendix_manipulation_example_list} for text descriptions, and Table~\ref{fig:characterization_table} for notation).}
    \label{fig:characterization_examples}
\end{figure*}

In this section, we address the need for a systematic understanding of the diversity of non-prehensile capabilities through a novel taxonomy.

Through considering 120 non-prehensile manipulations (Table~\ref{fig:appendix_manipulation_list}), we identified four key dimensions characterizing a manipulator at its contact surface with an object (Table~\ref{fig:characterization_table}, left). Two dimensions describe the surface's \emph{geometry:} its \emph{size} and its \emph{curvature}. Two describe its \emph{mechanical properties:} its \emph{friction}\footnote{
Since friction is a property of an interaction and not of a surface in isolation, this usage of the term is not strictly meaningful. We use the term as shorthand to refer to a ``typical'' coefficient of friction (CoF) for interactions between an end-effector surface and a target object. In this sense, we would say that materials like coarse sandpaper and silicone have high friction, while ice and Teflon have low friction.
If desired, values for this quantity could be standardized by choosing a standard target object for CoF measurements.} and its \emph{compliance}.

Because a contact surface can typically be considered as two-dimensional, each of the two geometric dimensions can be applied to the surface along two orthogonal axes. Thus in this scheme, a manipulator can be characterized by a set of six values.

In practice, this six-dimensional space can be discretized into a finite number of classes by binning values (Table~\ref{fig:characterization_table}, right). In the remainder of this paper, we choose a discretization in which size has three possible values (small, medium, large) and curvature, friction, and compliance each have two possible values (low, high). 
The result is 84 distinct classes (Fig.~\ref{fig:characterization_examples}).
The effective number of classes can be reduced further by observing that some groups are functionally equivalent (e.g., small--curved and small--flat would have the same effect in most applications).

Intuitively, a tool or multi-tool system that covers a wider spectrum of these dimensions will be able to effectively perform a larger range of non-prehensile tasks.
In an applied setting, a robot would likely only have access to a limited selection of these classes, chosen according to the tasks anticipated and constraints such as limited carrying capacity. The motivating goal would be to provide a minimal tool set containing enough versatility for a task environment while remaining of tractable size. 

Note also that some manipulators combine different possible capabilities depending on their orientation with respect to the object manipulated. For instance, with a knife, the sharp side can be used to cut an onion while the flat side can be used to transfer the chopped onion to a pan; with a spoon, the inside can be used to scoop up cookie dough while the outside can be used to press the dough into a mold; and with a pizza peel, the front lip can be used to slide under the pizza while the top can be used to carry it.
As a result, a smaller set of appropriately chosen tools, each providing multiple distinct interaction surfaces, can provide a larger set of effective options.

\section{Modular tool system\label{sec:toolsystem}}

\subsection{Design considerations}

In this section, we describe a prototype of a multi-tool manipulation system that can be used for non-prehensile manipulations. The goal of this system is to enable a robotic arm to autonomously use a palette of end-effectors with different properties, sampled from the characteristics discussed in the previous section, to execute a task. 

Our design takes the form of a tool changer system in which a one-armed robot with a two-finger gripper can autonomously select and load different passive end-effector modules onto a single common handle.
The system was designed according to the following considerations:
\emph{Low adoption barrier:} Compatible with industry-standard robotic arms and grippers.
\emph{Passive:} Completely passive mechanical system without the requirement for any additional degrees of actuation.
\emph{Low complexity:} No auxiliary or pass-through electronics, hydraulics, or pneumatics.
\emph{Modular:} Customizable system, flexible to extend with additional tools, to accommodate specialized needs that may arise.
\emph{Low volume, weight, cost:} System can be used without large overhead.
\emph{Reliability:} Components should be as simple and fail-safe as possible, with redundancy to mitigate complete system failure.

Alternative schemes considered included: (a) A single end-effector having multiple interaction surfaces with different properties. This approach would be limited in how many distinct surfaces could be accommodated in a compact design, or unwieldy if the design were built up to accommodate more surfaces (analogous to a daisy wheel or Selectric typewriter ball \citep{hickerson1959single}). (b) An actuated barrel containing multiple tools that could be selectively exposed \citep{pettinger2019passive}. This actuated approach would add mass, complexity, and bulk compared to passive alternatives. (c) A set of completely separate tools. This approach would require every tool to have its own handle,
whereas the modular approach in the design we chose allows features like specialized geometry to ensure a secure grip, and spring-loading to provide compliance as a built-in safety feature \citep{teeple2022multi}, to be incorporated into a single handle without requiring duplication of these overhead costs.

\subsection{Hardware}
\begin{figure*}[t]
    \centering
    \includegraphics[width=\textwidth]{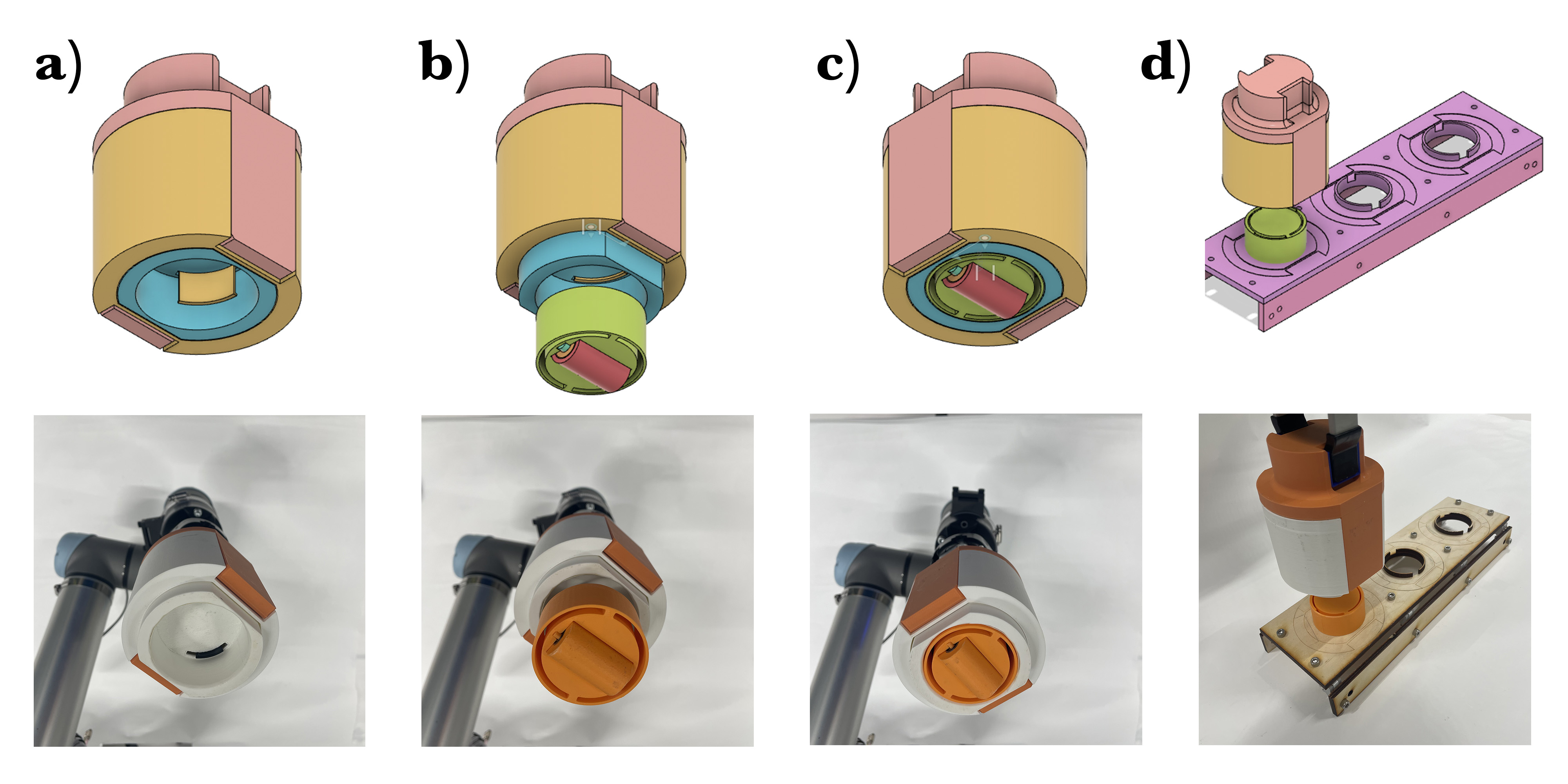}
    \caption{Selected prototype components designed and fabricated for the tool system (top: CAD renderings; bottom: 3D-printed prototypes). a) Tool holder, seen from below when held by gripper. A bayonet-mount retainer finger (small orange tab in rendering) is exposed inside a compliance unit (blue in rendering), which is connected to the outer housing through a spring. b) Tool holder aligned to roller insert (green in rendering) prior to mounting. c) Tool holder with roller insert mounted. d) Storage plate (pink in rendering) for storage and autonomous switching of different end-effector inserts (green), with tool holder aligned prior to mounting an insert.}
    \label{fig:design_overview}
\end{figure*}

The tool changer system comprises three principal components: (a) a tool holder, (b) various end-effector inserts with different properties, and (c) a storage plate to hold inserts while not in use (Fig.~\ref{fig:design_overview}). 

The cylindrical tool holder is the principal component of the modular tool system. It features a passive, rotary-style mounting system, a spring-loaded mechanism to add structural compliance, and a physical interface that can be easily used by different robotic grippers.

\begin{figure*}[t]
    \centering
    \includegraphics[width=\textwidth]{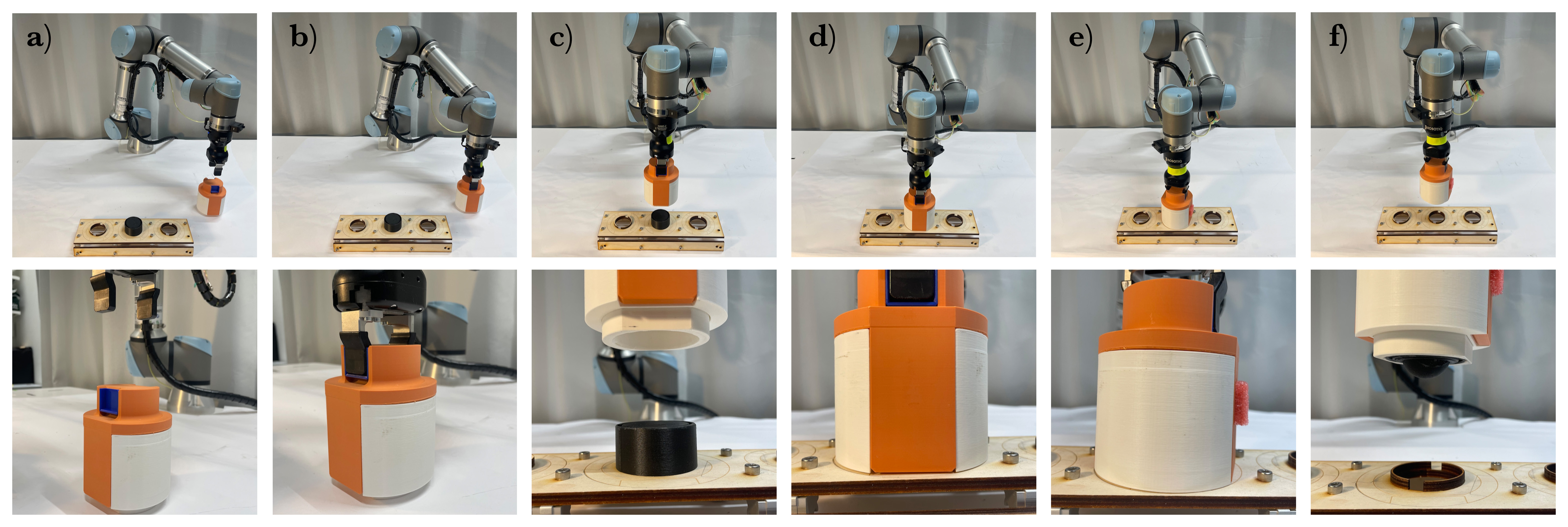}
    \caption{Mounting a tool insert  (top: wide view; bottom: closeup). a,~b) A standard prehensile gripper picks up the tool holder. c,~d) The holder is lowered onto the insert to be mounted, compressing the compliance unit and aligning the tool retainer fingers with chambers in the insert. e) Rotating the tool holder in place locks the insert. f) Lifting the holder uncompresses the compliance unit and exposes the insert for use.}
    \label{fig:design_tool_change}
\end{figure*}

The passive mounting system features a bayonet-style mechanism to attach different tool inserts, inspired by the mechanism found in interchangeable camera lenses. On the bottom of the tool holder, two arced retainer fingers extend into recesses present on the top side of the inserts (Fig.~\ref{fig:design_overview}A, B). When the arced fingers reach fully into the recesses of an insert stored on the tool changer plate, and the robot arm wrist rotates $100^\circ$ clockwise, they become locked in a narrow chamber in the tool insert, thus temporarily securing the tool to the tool holder. Reversing the operation by rotating the tool holder counter-clockwise while depressed into an open slot on the storage plate releases the insert again (Fig.~\ref{fig:design_tool_change}). 

To reduce complexity in the required control of the manipulation system \citep{teeple2022multi}, a compliant mechanism was added  to the tool holder. Compliance in the z-axis also helps prevent tool breakage. A top- and bottom-facing part inside the tool holder are separated by a spring. When the top of the tool holder is depressed, part of the energy applied to the tool compresses the spring.

Rather than a permanently mounted attachment, the tool changer system was designed as a device that a robot can pick up and put down again as needed, allowing it to switch back and forth quickly between prehensile and non-prehensile operations in a complex task.
The gripped interface of the tool holder features a high-friction surface texture and two tapered depressions that tightly fit the fingers of a gripper. When the gripper closes around the interface, the tool holder is kinematically locked to the gripper until released again. In our study, the interface was designed to fit a Hand-e by Robotiq, Inc.; the design can readily be adapted to other grippers.

\begin{figure*}[t]
    \centering
    \includegraphics[width=\textwidth]{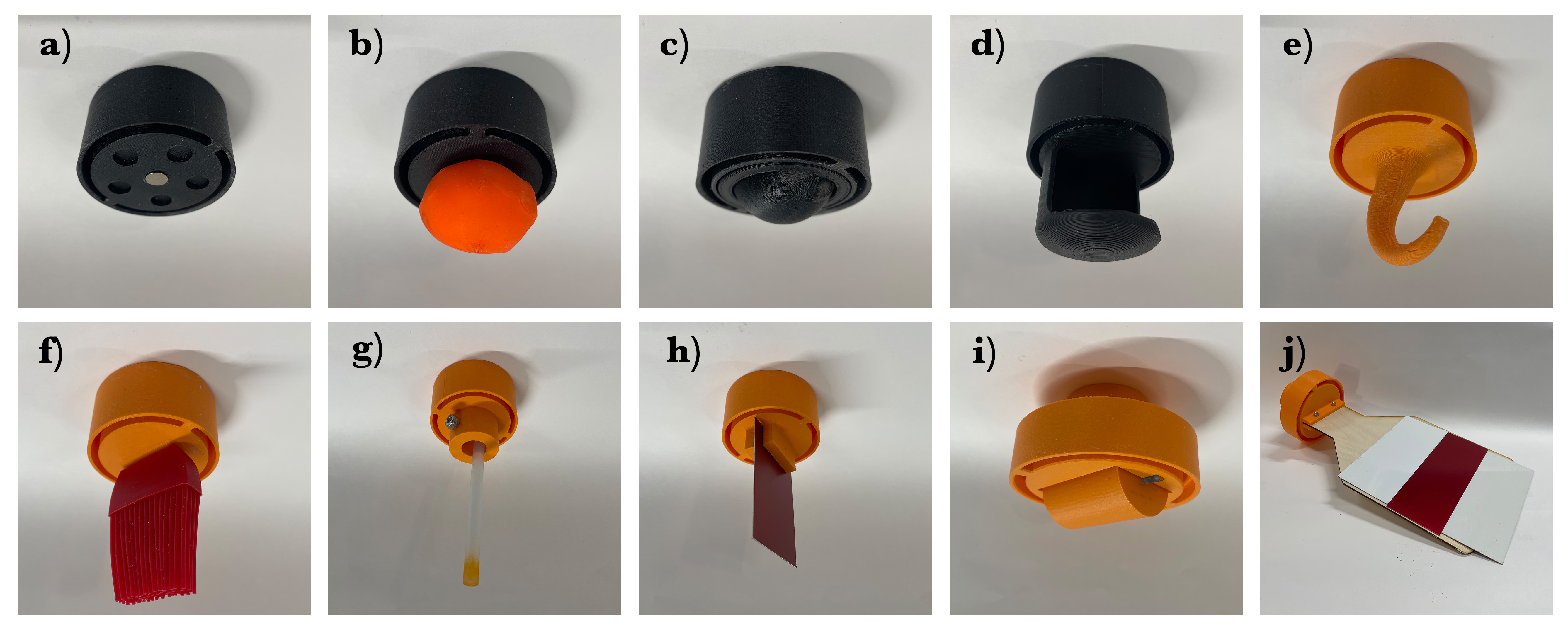}
    \caption{Tool inserts we fabricated for the two case study scenarios in \S\ref{sec:casestudies}: a) magnetic plate; b) soft silicone ball; c) hard plastic caster ball; d) spoon; e) hook; f) silicone brush; g) soft finger; h) burnishing tool; i) rolling pin; j) pizza peel.}
    \label{fig:scenario_tools}
\end{figure*}

Various tool inserts covering a wide range of the classes presented in the non-prehensile manipulation taxonomy were designed for use in potential application scenarios. In total, we designed and fabricated 10 tool insert prototypes for our two case study scenarios (Table~\ref{fig:scenario_tools}).

A simple storage plate was designed to store the inserts and prevent them from turning while the tool holder locks/unlocks them.

The tool holder and inserts were fabricated from 0.4mm Prusament PLA on a Prusa MK4S+ 3D-printer. The tool holder incorporated a music-wire spring (spring force: 22 lbs./in.); the inserts incorporated other off-the-shelf materials, described in more detail in the case studies below. The tool holder was 85mm $\times$ 95mm $\times$ 110mm and 240g; the inserts are at least 55mm $\times$ 55mm $\times$ 27.5mm and 25g.
The plate was fabricated from laser-cut 1/4'' plywood plates and standard aluminum extrusion corner braces.

\subsection{Reliability}

To quantify the reliability of the tool change operation, a repeatability study was performed. A robotic arm (UR5e by Universal Robots, Inc.), equipped with a Robotiq, Inc. Hand-e wrist module and parallel jaw gripper with silicone-covered fingertips, autonomously mounted and unmounted a tool insert repeatedly over n=200 trials, while a human observer visually verified whether each attempt was successful. 
In total, 199/200 attempts were fully successful. In the exception, the tool was not perfectly aligned with the arced retainers on the tool changer plate at first, but slid into place during the course of the unmounting operation.

\section{Case study scenarios\label{sec:casestudies}}

Here, we present two example case studies to demonstrate the use of the manipulation system in relevant scenarios, involving tasks that require a variety of non-prehensile manipulations.
Note that the emphasis of this work is on the physical affordances of the tools and the utility these provide, rather than on sensing and control for task execution in unknown environments; therefore, while the robot performed the case study tasks autonomously without human intervention, it did so open-loop.

\subsection{Scenario 1: Leak patching in a space habitat}

In the first scenario, a robot was tasked with the repair of a small leak in an interior wall of a space habitat. The task involves localizing the leak using an acoustic detector, and then applying an adhesive patch over the hole to stop air from escaping. 
Repair of such holes is an infrequent but critical task during crewed space missions \citep{heiken1991lunar, suggs2015results}, and a priority identified for autonomous robotic operation during future missions, which will include extended uncrewed periods \citep{laske2020eci}.

Component fabrication, tool selection, and repair procedures were based on a patch kit developed for the International Space Station for this purpose \citep{nasaJSC28533E}.
Self-adhesive, metallic patches were designed to match the thickness, pliability, and texture of those in the kit \citep{coronado1987wall}. The 2''-wide hexagonal patches were made from a multi-surface-adherent textile tape bottom, a pliable middle
layer from 3M VHB, a thermally-conducting aluminum tape top layer, and a standard ferromagnetic, zinc-plated washer. In order to permit one-handed operation of a patch, a robot factors design approach \citep{melenbrink2022robot} was taken: a novel dispensing cartridge removed the adhesive backing from a patch when it was withdrawn; and a special-purpose tool insert was designed for this scenario, with a flat surface with embedded magnets that permitted lifting the patch. We also fabricated an acoustic detector to localize the leak, containing an electret microphone amplifier module, a wireless ESP-32 \citep{xiaoesp32}, and a generic 2000 mAh rechargeable battery. For expediency, we attached the detector directly to the tool changer.

\begin{figure*}[t]
    \centering
    \includegraphics[width=\textwidth]{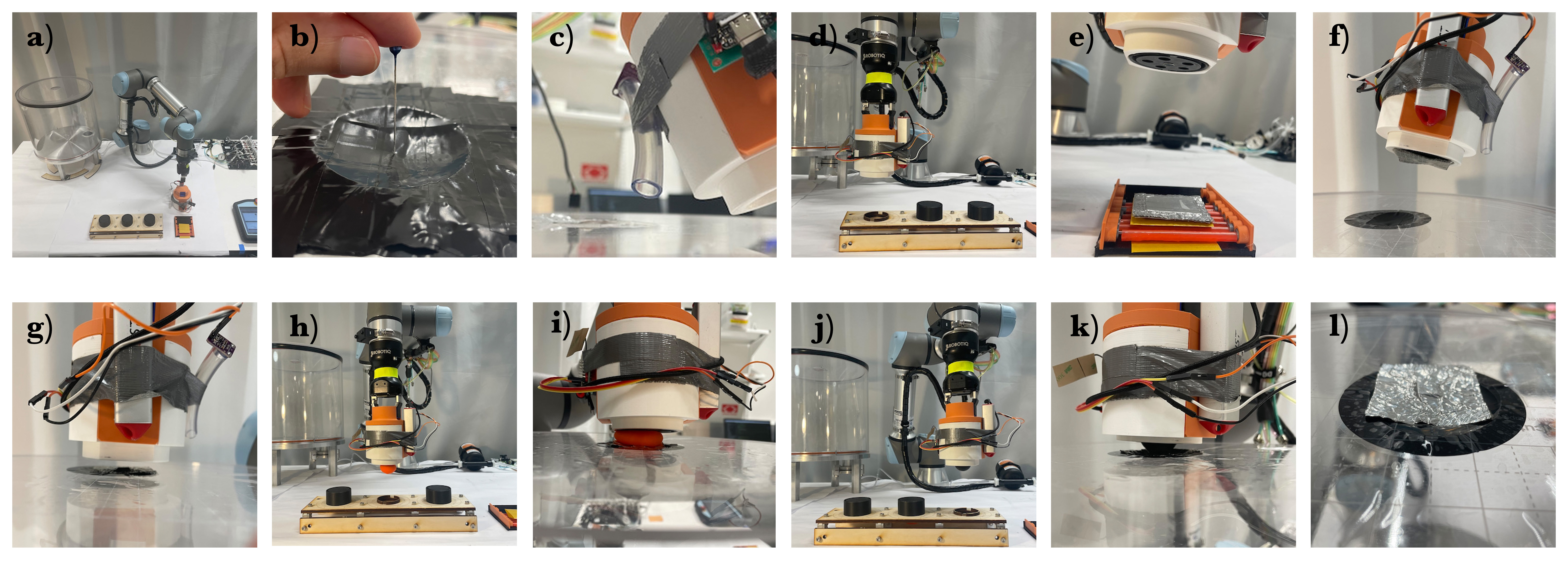}
    \caption{Scenario 1: Operations sequence. a) Work area in initial state; b) hole in soft cover over vacuum chamber lid; c) leak finder tool in use; d) magnetic tool mounted; e) magnetic tool picking up adhesive patch from cartridge; f) approaching the leak; g) initial patch placement; h) soft silicone tool mounted; i) soft silicone tool in use; j) hard plastic caster tool mounted; k) hard plastic caster tool in use; l) result of patch operation. See also Video 1.}
    \label{fig:scenario_1_tool_change}
\end{figure*}

The repair procedure consisted of four steps (Fig.~\ref{fig:scenario_1_tool_change}): (1) Perform a raster scan of the candidate leak area with the acoustic detector to localize the leak. (2) Use a magnetic tool (class M–M–cf) to remove a patch from the dispenser, and press it down at the center of the leak. (3) Use a soft tool (class M$\cap$M$\cap$CF) to press the patch down more securely. (4) Use a caster ball tool (class S$\cap$S$\cap$cf) to press down around the perimeter of the patch to finalize the repair.

For physical simulation of the leaking habitat, we used an off-the-shelf acrylic vacuum chamber by Abbess Instruments and Systems, Inc.\ and fabricated a new lid with a set of inserts enabling holes of various sizes. The interior of the chamber corresponded to the exterior of a space habitat (i.e., the vacuum of outer space).
The vacuum chamber was continuously evacuated at maximum flow rate using house vacuum.
When a leak was initiated, the low pressure inside the chamber gradually increased toward atmospheric pressure, so that the pressure difference between the chamber exterior and interior corresponded to the pressure difference that would exist between a habitat interior and the external vacuum.
We tested patching leaks in both rigid and soft wall panels. For rigid wall panels, 2'' diameter polymethyl methacrylate (PMMA) discs with holes of different sizes were used as inserts in the vacuum chamber lid. For soft wall panels, 10'' diameter circular sheets of polyvinyl chloride (PVC) were adhered onto the vacuum chamber lid, covering the opening, and pierced using a 0.5mm diameter dressmaker's pin at the start of the experiment.

\begin{figure}[t]
    \centering
    \includegraphics[width=\columnwidth,height=\textheight,keepaspectratio]{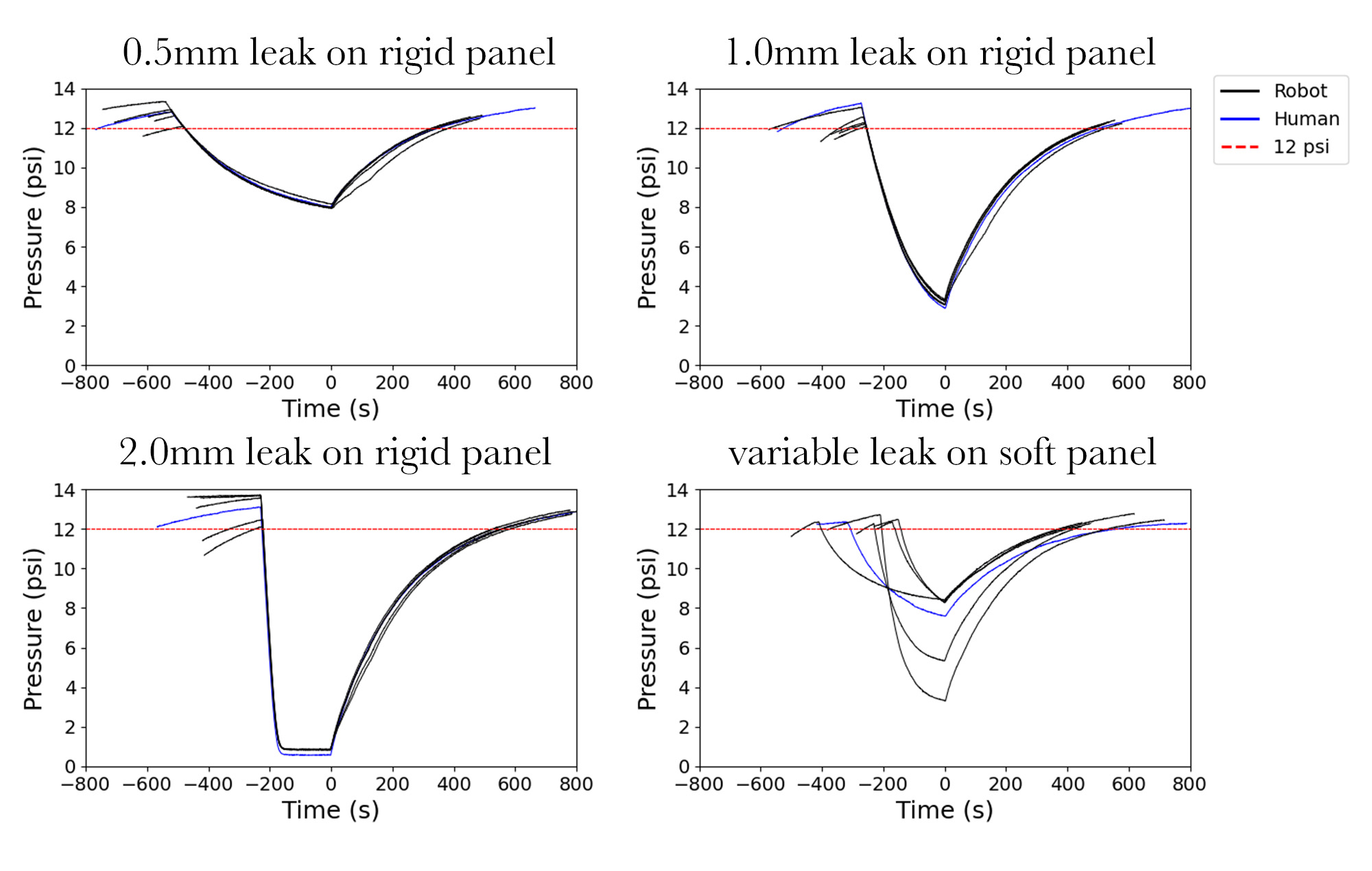}
    \caption{Results from leak patching in scenario 1. Absolute pressure difference between atmosphere and inside of vacuum chamber for five repeated experiments per leak size, with four leak sizes. Observations from experiments are aligned such that initial patch placement occurs at t=0 min. 12 psi indicates normal pressure difference. Leak repair was initiated when the pressure difference dropped below 12 psi for a hard wall panel, and at 9 psi for a soft wall panel.}
    \label{fig:scenario_1_results}
\end{figure}

We evaluated the success of a patching attempt according to the recovery of the vacuum within the chamber, quantified as the pressure difference between the interior and the exterior of the chamber. For our experiment, nominal pressure was defined as a pressure difference above 12 psi. For rigid panels, patching was autonomously initiated when pressure dropped below 12 psi; for soft panels, when it dropped below 9 psi. The success criterion was for pressure to return above 12 psi after patching.
A total of 20 trials were conducted by the robot using the described tool system, five each for rigid wall panels with holes of size \{0.5mm, 1.0mm, 2.0mm\} and soft wall panels. For comparison, one trial for each condition was conducted by a human operator.

Results showed that the robot was able to fix the leak in all 20 trials, restoring high vacuum fidelity (Fig.~\ref{fig:scenario_1_results}).
We quantitatively compared the robot's performance to that of a human operator performing the same task, using the area under the curve (AUC) of the pressure recovery as a metric. Our null hypothesis is that the robot AUCs are normally-distributed samples from a population with a true mean of the human operator AUC; if the human operator is consistently better than the robot, this hypothesis would be rejected. One-sample t-tests for the four conditions (leak sizes of 0.5mm, 1.0mm, 2.0mm, and soft leaks) gave p-values of 0.1642, 0.3352, 0.4753, and 0.8559, respectively. Thus we find that the robot's performance was comparable to that of a human operator in all conditions tested.

\subsection{Scenario 2: Personal pizza making}

In this second scenario, a robot was tasked with the preparation of personal pizzas for later baking, starting from a (prepared) set of ingredients. We chose this task as one that requires a range of different prehensile and non-prehensile manipulations, and one that serves as an exemplar of kitchen and food preparation tasks commonly envisioned for future household robots.

\begin{figure*}[t]
    \centering
    \includegraphics[width=\textwidth,height=\textheight,keepaspectratio]{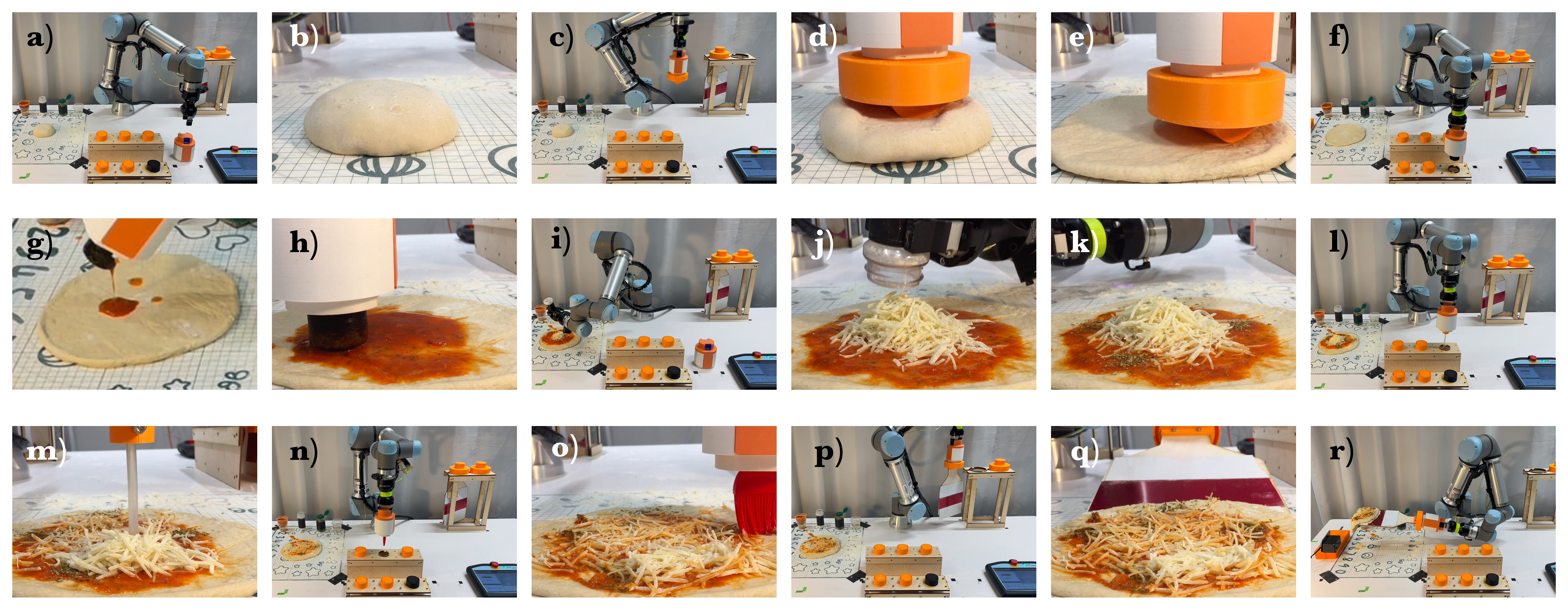}
    \caption{Scenario 2: Operations sequence. a) Work area in initial state; b) pizza dough in initial state; c) rolling pin tool mounted; d, e) flattening pizza dough using rolling pin tool; f) spoon tool mounted; g) dispensing tomato sauce using spoon tool; h) spreading tomato sauce using underside of spoon tool; i) unmounted tool holder to use gripper for cheese and spices; j) dumping out cheese from container; k) dispensing spices from canister; l) tool holder and finger tool remounted; m) spreading cheese using finger tool; n) silicone brush tool mounted; o) covering crust in olive oil using brush; p) pizza peel mounted; q) using pizza peel to scrape pizza from work surface; r) using pizza peel to lift pizza dough. See also Video 2.}
    \label{fig:scenario_2_tool_change}
\end{figure*}

\begin{figure}[t]
    \centering
    \includegraphics[width=\columnwidth,height=\textheight,keepaspectratio]{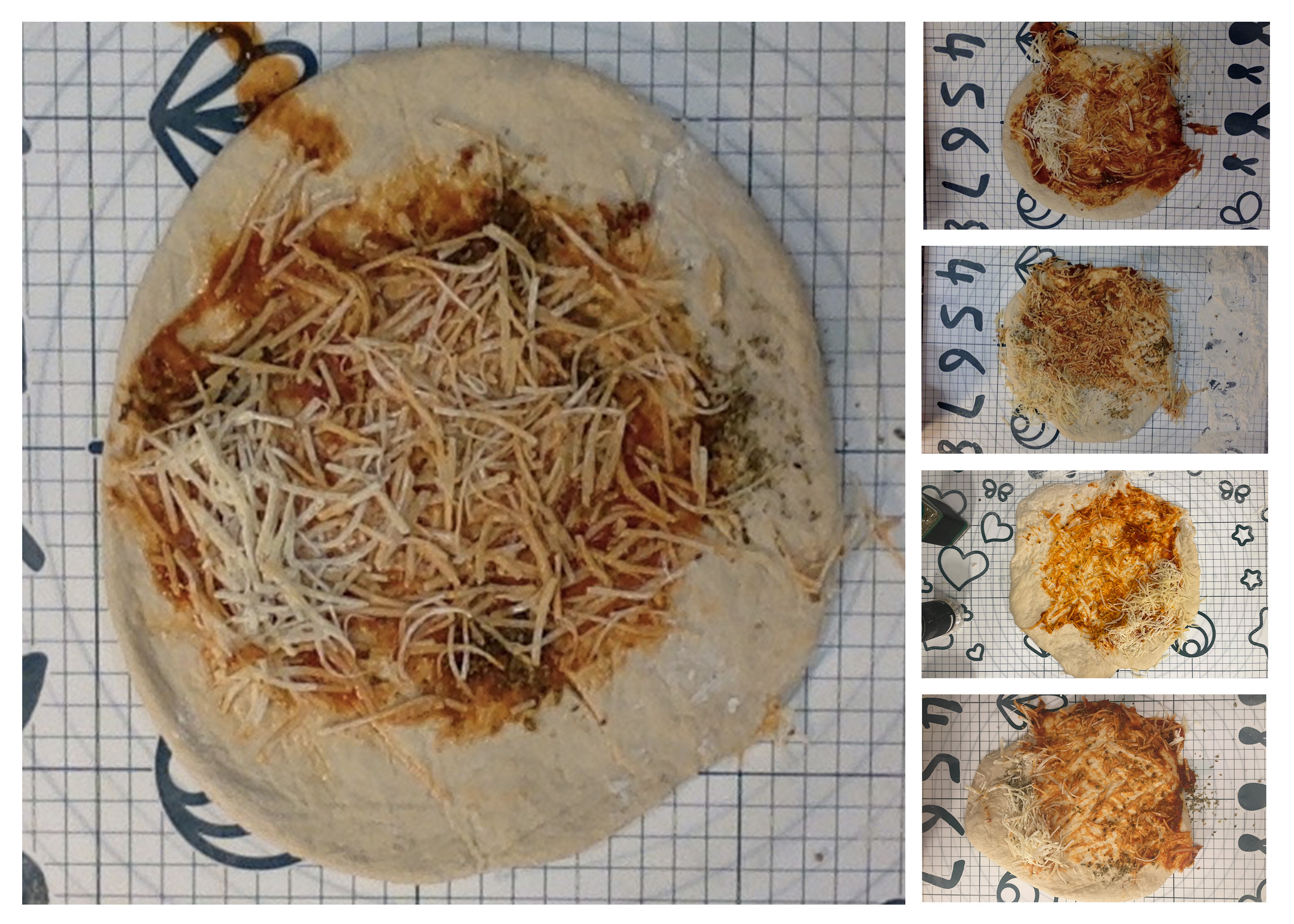}
    \caption{Scenario 2: Five personal pizzas prepared in succession.}
    \label{fig:scenario_2_results}
\end{figure}

Tool inserts were chosen based on the non-prehensile manipulation classes of tools found in pizza-making recipes and videos \citep{artusi_1891,hazan_1973,bbc_pizza}. 
The initial condition consisted of a ball of pizza dough in the center of a floured 24'' $\times$ 16'' silicone baking mat; small containers of tomato sauce, shredded cheese, and olive oil; and a seasoning jar of thyme, all placed in known locations.

The pizza preparation procedure consisted of seven steps (Fig.~\ref{fig:scenario_2_tool_change}): (1) Use a pizza roller tool (class L–M$\cap$cf) to roll out the dough to a suitable thickness. (2) Use a spoon insert (class M$\cap$M$\cap$cf) to scoop sauce onto the dough (using the upper surface of the insert) and spread it evenly (using the lower surface of the insert). (3) Pick up the bowl of cheese and dump it onto the pizza (a prehensile operation). (4) Pick up the jar of thyme and sprinkle it onto the pizza (a prehensile operation). (5) Use a finger-like insert (class S–S–CF) to distribute the cheese over the sauce. (6) Use a silicone brush insert (class M–S–CF) to coat the rim of the pizza with olive oil. (7) Use a pizza peel insert to slide under the pizza (class L–S$\cap$cf) and lift it from the mat (class L–L–Cf).

To demonstrate a consistent ability to perform this task, the robot prepared five pizzas in a row (Fig.~\ref{fig:scenario_2_results}).

\section{Discussion}

The modular manipulation system presented here is distinct from existing tool-changer systems in that it provides a completely passive system with limited complexity to perform a 
range of non-prehensile manipulations. As it can temporarily be picked up by a prehensile manipulator (e.g., a robotic gripper), the system extends the affordances of that manipulator with minimal overhead. In contrast, other tool changers used for non-prehensile manipulations are permanently mounted to a robotic arm and show higher complexity: Many are active \citep{ryuh2006automatic, gyimothy2011experimental, smith2020modular, meghdari2000design} or have pass-through couplings \citep{li2022modular}. The few passive tool changers have complex mechanical locking mechanisms not suitable for autonomous operation by a one-armed robot, in the absence of an external auxiliary actuated system \citep{berenstein2018open, pettinger2019passive}.

A principal benefit of the tool-changer-based framework for our manipulator is the fact that functionality shared across tools can be abstracted to the tool changer level, while functionality pertaining to a single tool can be specified on the tool insert level. This abstraction reduces overall weight, volume, and complexity. For instance, the spring-loading mechanism only had to be integrated once as part of the tool changer, and a reinforced gripper interface on the tool changer instead of on separate tools yields a stronger handle without adding extra weight or volume.

The separation of the tool changer and tool inserts also makes the modular tool system suitable to collaborative robotics settings. By sharing the same set of tool inserts between multiple robots, each robot only requires its own tool changer. Furthermore, as the tool changer is designed to temporarily be picked up by a general-purpose gripper, it could be shared between different robotic systems and arms in the same task environment.

The taxonomy was found to be a valuable mental model for tool design and choice. 
In the context of a given task, the taxonomy helped in choosing scenario-specific tool palettes with clearly-defined capabilities. By mapping desired task affordances to tool inserts, an appropriate subset of tools can be chosen that fulfills task demands without extraneous redundancy. This consideration is especially relevant in weight- and volume-constrained scenarios, as with space robotics.

The taxonomy scheme stands out as a framework that does not depend on the geometry or functions of the human hand, but rather focuses on the physical properties of any given end-effector. In comparison, many prior taxonomies \citep{schlesinger1919mechanische, cutkosky1989grasp, feix2015grasp} only apply to anthropomorphic hands. Focusing on the end-effector properties means our characterization scheme can also be applied to other types of manipulators, including robotic grippers and human prostheses.

While we found the taxonomy as presented here to be useful for a range of operations and needs, it is not expected to be exhaustive for all possible scenarios or end-effectors. One relevant consideration is that of manipulators that present multiple simultaneous contact points or surfaces (e.g., fork, rake, brush, etc.). Such end-effectors could be treated as a blended continuous surface between these points, or as a combination of multiple distinct surfaces with individual classes. More broadly, for specific scenarios with more granular demands, custom classification schemes could build on this one and extend it with further classes as required.

We highlight three areas for future extensions to this work:

\emph{(1) ``Non-prehensile grasping''.} The magnetic tool insert used in the patching scenario, which enabled lifting the patch without grasping it, points to the possibility of other modes for picking up objects using a passive, non-prehensile manipulator. In particular, adhesive surfaces on a tool insert can similarly be used to lift small objects. (Adhesion may be informally accommodated within our taxonomy by considering it as a limiting case of high friction.)

\begin{figure}[tp]
    \centering
    \includegraphics[width=\columnwidth,height=\textheight,keepaspectratio]{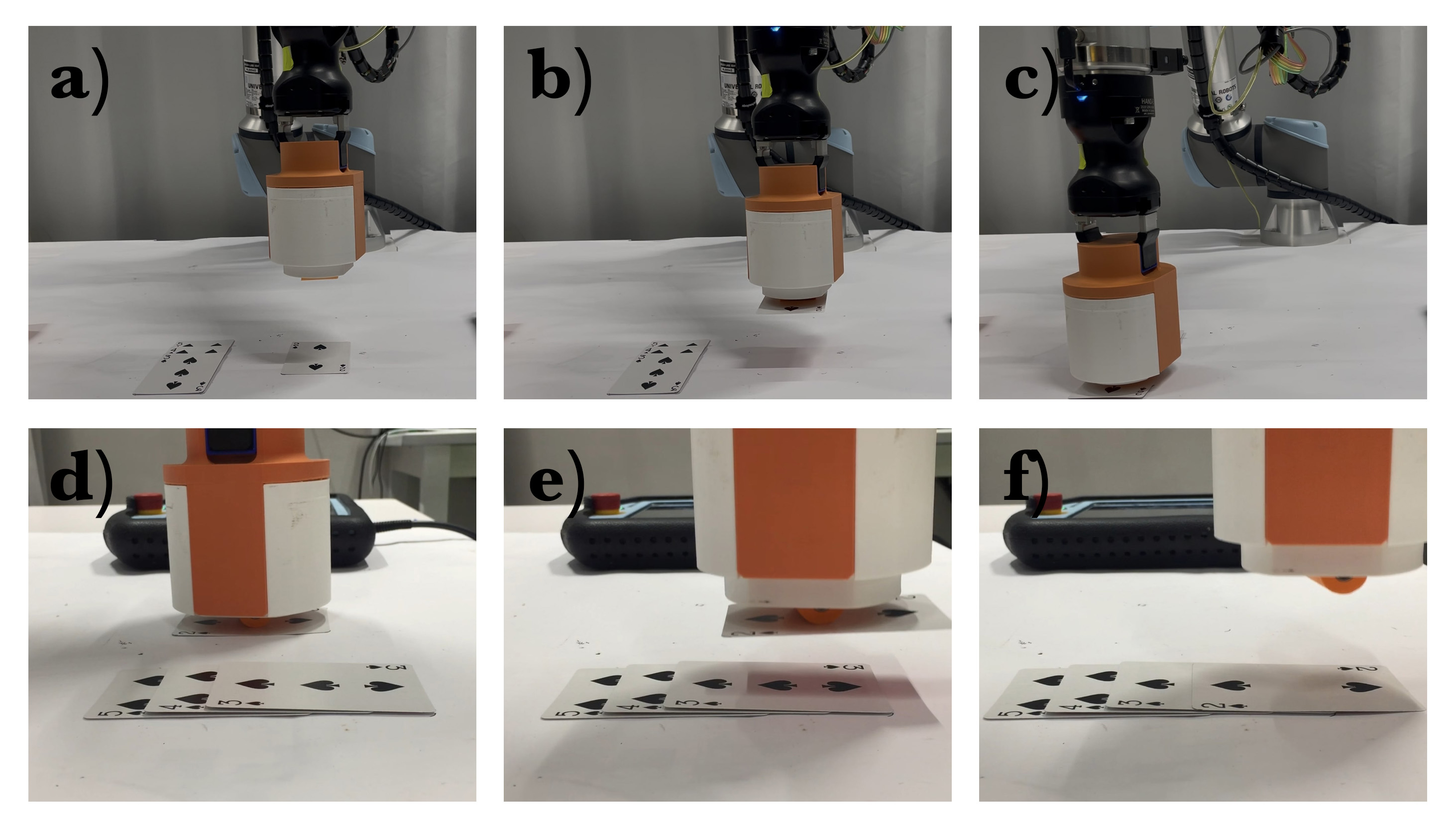}
    \caption{Demonstration of ``non-prehensile grasping''. a) A tool insert provides as an end-effector a rolling cylinder, half of whose surface is coated with an adhesive layer. Initially, the adhesive side is contained within the insert, and the smooth side is exposed. b) Rolling the cylinder over the playing card brings the adhesive into contact with the card and binds it to the tool. c–f) Rolling in reverse releases the card.}
    \label{fig:non-prehensile_grasping}
\end{figure}

We explored this type of manipulation by creating an additional tool insert, consisting of a rolling cylinder with an adhesive layer over half its surface area and a smooth surface on the other half. (A useful analogy might be a lint roller, sticky on only one side.) The insert is used by putting it in contact with a substrate smooth side down, and rolling it halfway so that the sticky side comes into contact with the object to be lifted; reversing the operation releases the object again (Fig.~\ref{fig:non-prehensile_grasping}, Video 3).

Note that this approach can be particularly useful for thin, light objects (paper, light cloth, etc.), which by the same token are particularly challenging for traditional rigid grippers \citep{teeple2022multi}. Thus such a non-prehensile tool is again complementary to a standard prehensile gripper.

\emph{(2) Prehensile finger surfaces.} The same factors that make the classification scheme useful for non-prehensile manipulation could also be applied to prehensile manipulators. A robot could don interchangeable covers for individual fingers 
 (``thimbles''), or for the entire gripper (``gloves''), according to the desired manipulation properties for a task. For instance, thimbles could provide temporary fingernails to help with scratching targets or lifting thin objects, a nitrile glove could be used to improve the hold of an irregularly-shaped object, or a cotton glove could be used to prevent scratches when handling delicate glassware.

\emph{(3) Closed-loop autonomy.} As noted above, the case studies explored here used open-loop control and pre-programmed tool paths. An important topic for future work will be the use of such a system by an intelligent robot: choosing the most appropriate tool according to the manipulation demands of a given operation, and using that tool to perform the operation effectively. The taxonomy may be a useful framework for such a robot to employ in selecting tools, as we found it to be for our hand-designed task execution.

\subsection*{Funding acknowledgment}

This work was supported by a Space Technology Research Institutes grant (number 80NSSC19K1076) from NASA's Space Technology Research Grants Program.

\bibliography{references}

\appendix
\clearpage
\renewcommand{\thepage}{S\arabic{page}}
\renewcommand{\thesection}{S\arabic{section}}
\renewcommand{\thetable}{S\arabic{table}}
\renewcommand{\thefigure}{S\arabic{figure}}
\setcounter{figure}{0}    
\setcounter{page}{1}

\begin{figure*}[t]
    \centering
    \includegraphics[width=\textwidth, height=0.9\textheight, keepaspectratio]{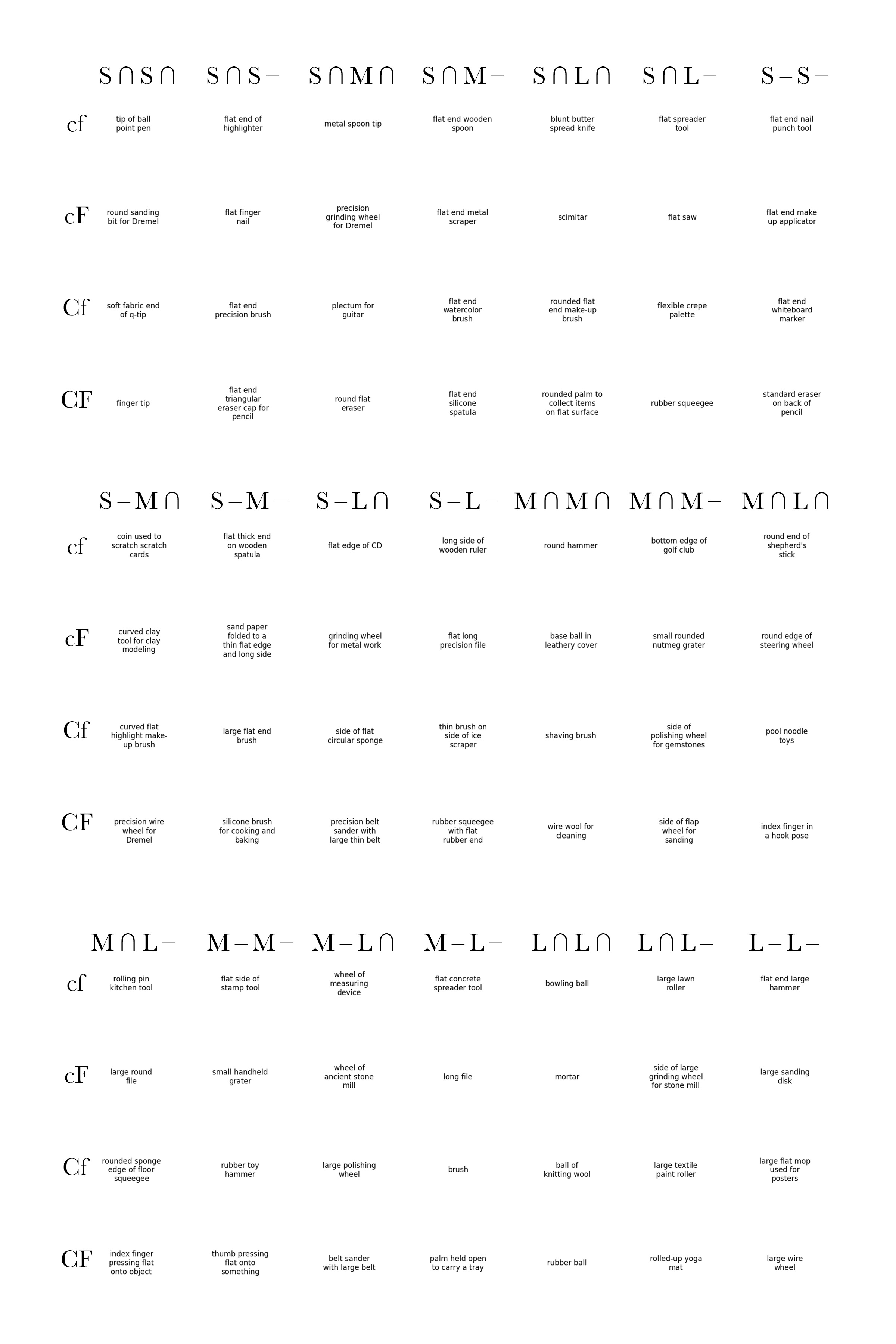}
    \caption{Descriptions of the examples presented in Fig.~{\ref{fig:characterization_examples}}}
    \label{fig:appendix_manipulation_example_list}
\end{figure*}

\begin{table*}[t]
    \centering
    \caption{Non-prehensile manipulations considered during development of taxonomy in Table \ref{fig:characterization_table}}
    \label{fig:appendix_manipulation_list}

    \fbox{
        \begin{varwidth}{\textwidth}
        {\tiny
        \begin{multicols}{2}
            \begin{itemize}[noitemsep]
                \item using thumbs to massage neck
\item using fine paint brush to draw oil painting
\item using coarse paint brush to draw acrylic painting
\item using painting roll to paint drywall
\item using fine silicone applicator for make up
\item using shaving brush to apply shaving cream
\item using fist to punch someone/something
\item using edge of flat hand for karate
\item using finger nail to scratch dirt from a surface
\item using pen to sign a document
\item using letter knife to open a letter
\item using flat hammer to drive a nail into a wall
\item using box cutter to cut cardboard
\item using wooden spatula to press tongue down
\item using wooden spatula to stir a liquid
\item using finger to dip into a sauce to test its taste
\item using finger tip to press a steak to feel its tenderness
\item using flat hand with a cloth to clean a window
\item using flat hand to do a face palm
\item using flat hand to cover ears from sound
\item using finger to to cover ear from sound
\item using heel of hand to knead pizza dough
\item using flat hand to give a high five
\item using finger tip and nail to pick up grain of rice
\item using broom to clear leaves from a driveway
\item using tooth brush to brush teeth
\item using sponge to clean dishes
\item using knife to cut vegetables
\item using knife to pick up chopped vegetables
\item using burger press to make a burger patty
\item using file to file sharp metal edge
\item using deburring tool to remove plastic burrs
\item using sand paper to sand wood
\item using chesse grater to grate cheese
\item using thumbs to massage neck
\item using heels of hand to massage back
\item using flat hand to feel belly
\item using finger tip to tap a touch screen
\item using stylus to write on a touch screen
\item using saw to cut wood
\item using tooth pick to remove food residue
\item using dental floss to remove plaque
\item using interdental brushes to remove plaque
\item using eyeliner to apply make up
\item using sponge to clean in bathtub
\item using wire brush to remove rust
\item using grinding wheel to polish metal
\item using table saw to cut plastic
\item using butter knife to cut butter
\item using butter knife to spread butter
\item using spoon to scoop up liquid
\item using spoon to spread liquid
\item using drill bit to drill a hole
\item using buffing machine to buff hardwood floor
\item using mop to clean floor
\item using duster to remove dust
\item using lint roller to remove lint
\item using spreading tool for wet concrete
\item using large rubber squeegee to clean floor
\item using small rubber squeegee to clean black board
\item using silicone brush to apply egg to pastries
\item using baking palette to flip a crepe
\item using sharpie to mark object
\item using highlighter to highlight text
\item using garlic press to prepare minced garlic
\item using side of finger to flip a light switch
\item using finger tip to press a button
\item using moistened finger tip to flip a newspaper page
\item using mallet to press part into place
\item using fabric hammer to sound a gong
\item using spatula to flip a steak
\item using finger tip to feel temperature of water in a bath tub
\item using printing press to print a page
\item using stamp to stamp a letter
\item using metal stamp to seal a letter with wax
\item using elbow to open door without touching handle with hand
\item using sides of foot to hold door open
\item using heel of foot to crack a walnut
\item using whole body for wrestling
\item using side of arm for karate
\item using shoulder blade to roll over
\item using finger in hook pose to grab an old-style plug from kitchen sink
\item using hand in hook pose to grab shopping bag
\item using shoulder to hold hand bag
\item using bottle brush to clean bottles
\item using wire cutter to cut wires
\item using crow bar to pry objects apart
\item using finger nail to select page to open in a closed book
\item using flat hand to roll rolls of dough
\item using safety razor to shave
\item using metal scraper to scrape grime from stone kitchen surface
\item using metal plane tool to trim wood
\item using dressmaker's pin to pop a bubble
\item using a comb to comb hair
\item using a hair brush to brush hair
\item using scalp massager to massage scalp
\item using inner side of foot to pass a soccer ball
\item using head to head a soccer ball
\item using knees to juggle with a soccer ball
\item using head to spin a basket ball
\item using base ball bat to hit a base ball
\item using ice skates to skate on ice
\item using roller blades to skate on street
\item using bar jigger to mash mocktail ingredients
\item using lemon squeeze to juice a lemon
\item using meat pounding tool to prepare schnitzel
\item using ladle to scoop soup
\item using nutmeg grater to grate fresh nutmeg
\item using wire sponge to remove grime from pan
\item using silicone spatula to spread dough
\item using dough knife to part dough
\item using flour brush to remove flour from work surface
\item using long spoon to mix mocktail ingredients
\item using meat grinder to prepare minced meat
\item using edge of ruler to fold paper
\item using iron to iron a shirt
\item using hair iron to straighten hair
\item using curl inserts to curl hair
\item using silicone roll to apply ink to linoleum cut
\item using wooden roll to spread cookie cutting dough
            \end{itemize}
        \end{multicols}
        }
        \end{varwidth}
    }
    
\end{table*}

\end{document}